\documentclass[letterpaper]{article} 
\usepackage{aaai24}  
\usepackage{times}  
\usepackage{helvet}  
\usepackage{courier}  
\usepackage[hyphens]{url}  
\usepackage{graphicx} 
\usepackage{natbib}  
\usepackage{caption} 
\usepackage{algorithm}
\usepackage{algorithmic}

\usepackage{amsmath}
\usepackage{graphicx}
\usepackage{subfig}
\usepackage{multirow}
\usepackage{pgfplots}
\usepackage{booktabs}
\usepackage{bm}
\usepackage{amsfonts,amssymb}

\usepackage{newfloat}
\usepackage{listings}
\DeclareCaptionStyle{ruled}{labelfont=normalfont,labelsep=colon,strut=off} 
\floatstyle{ruled}
\newfloat{listing}{tb}{lst}{}
\floatname{listing}{Listing}
\title{Every Node is Different: Dynamically Fusing Self-Supervised Tasks \\ for Attributed Graph Clustering}
\author{
    Pengfei Zhu, Qian Wang, Yu Wang\thanks{Corresponding author}, Jialu Li, Qinghua Hu
}
\affiliations{
    College of Intelligence and Computing, Tianjin University, Tianjin, China \\
    \{zhupengfei, wang\_qian0806, wang.yu, jialuli, huqinghua\}@tju.edu.cn
}

\usepackage{bibentry}

\begin{document}

\maketitle

\begin{abstract}
Attributed graph clustering is an unsupervised task that partitions nodes into different groups. Self-supervised learning (SSL) shows great potential in handling this task, and some recent studies simultaneously learn multiple SSL tasks to further boost performance. 
Currently, different SSL tasks are assigned the same set of weights for all graph nodes.  
However, we observe that some graph nodes whose neighbors are in different groups require significantly different emphases on SSL tasks. 
In this paper, we propose to dynamically learn the weights of SSL tasks for different nodes and fuse the embeddings learned from different SSL tasks to boost performance. 
We design an innovative graph clustering approach, namely Dynamically Fusing Self-Supervised Learning (DyFSS). 
Specifically, DyFSS fuses features extracted from diverse SSL tasks using distinct weights derived from a gating network. 
To effectively learn the gating network, we design a dual-level self-supervised strategy that incorporates pseudo labels and the graph structure. 
Extensive experiments on five datasets show that DyFSS outperforms the state-of-the-art multi-task SSL methods by up to 8.66\% on the accuracy metric. The code of DyFSS is available at: https://github.com/q086/DyFSS.
\end{abstract}

\section{Introduction} 

Attributed Graph Clustering (AGC) aims to partition the graph nodes into several groups without explicit supervision. A key challenge of AGC is the lack of proper supervision~\cite{CDRS, RGAE}. Self-supervised learning (SSL) finds supervision from the data and has been shown to be greatly helpful for obtaining good performance~ \cite{SSLGNNs:Survey}. 

\begin{figure*}[!ht]
    \centering
    \subfloat[]
    {
        \label{graph-demo}
        \includegraphics[width=0.2\linewidth]{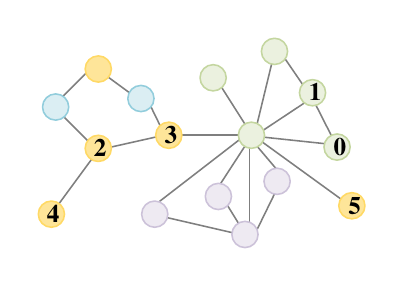}
    }
    \subfloat[]
    {
        \label{par-sim}
        \includegraphics[width=0.2\linewidth]{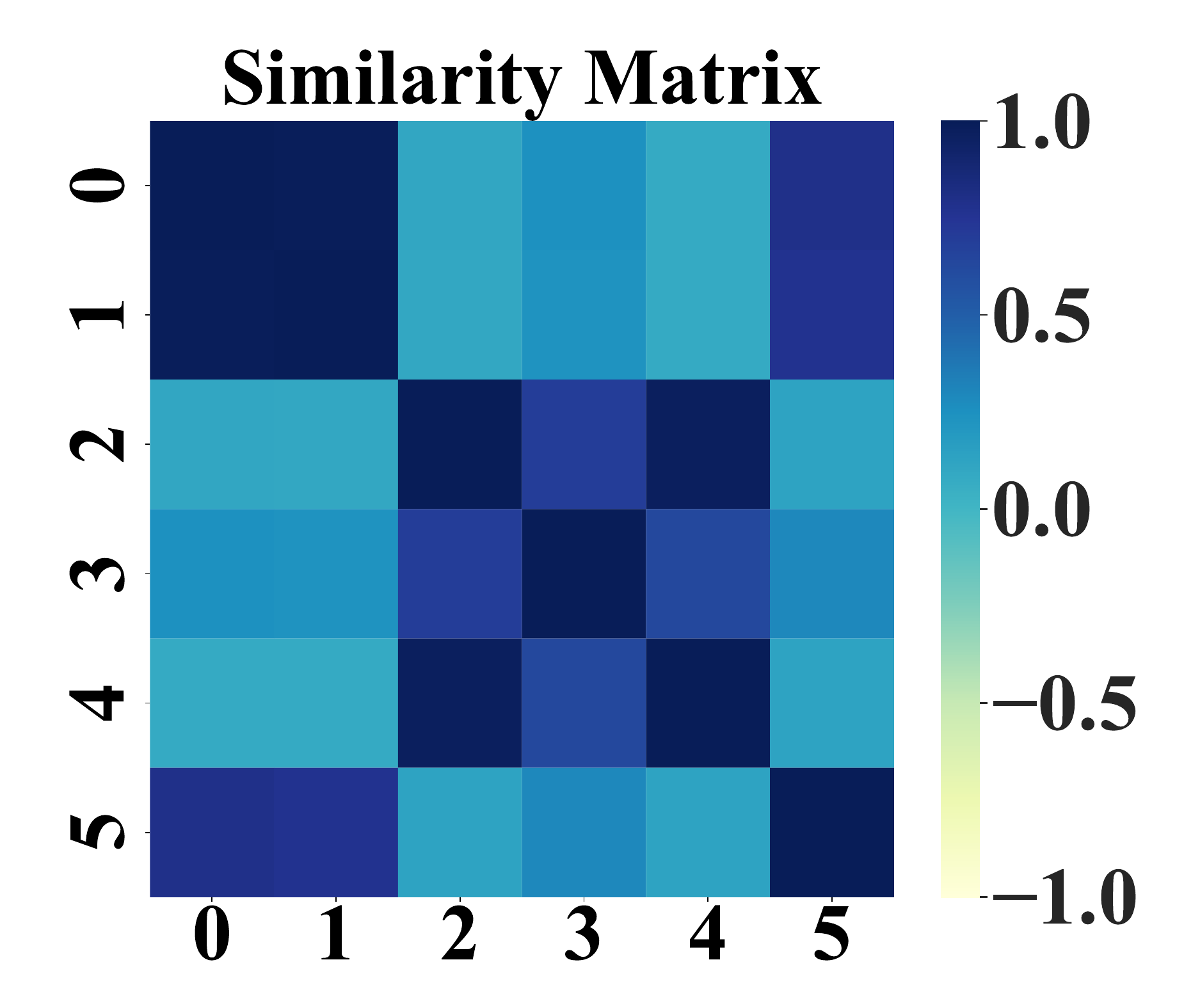}
    }
    \subfloat[]
    {
        \label{clu-sim}
        \includegraphics[width=0.2\linewidth]{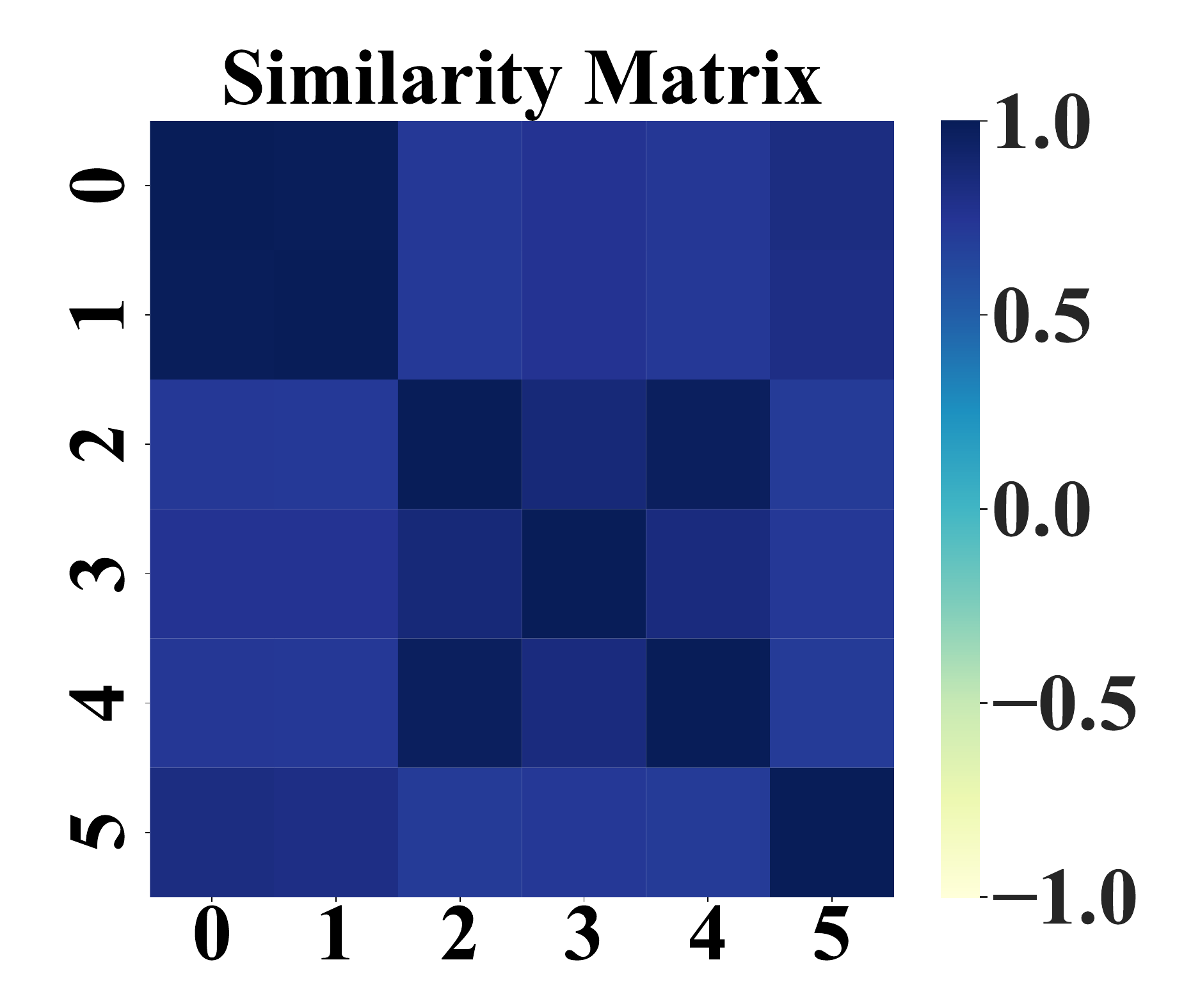}
    }
     \subfloat[]
    {
        \label{AutoSSL-sim}
        \includegraphics[width=0.2\linewidth]{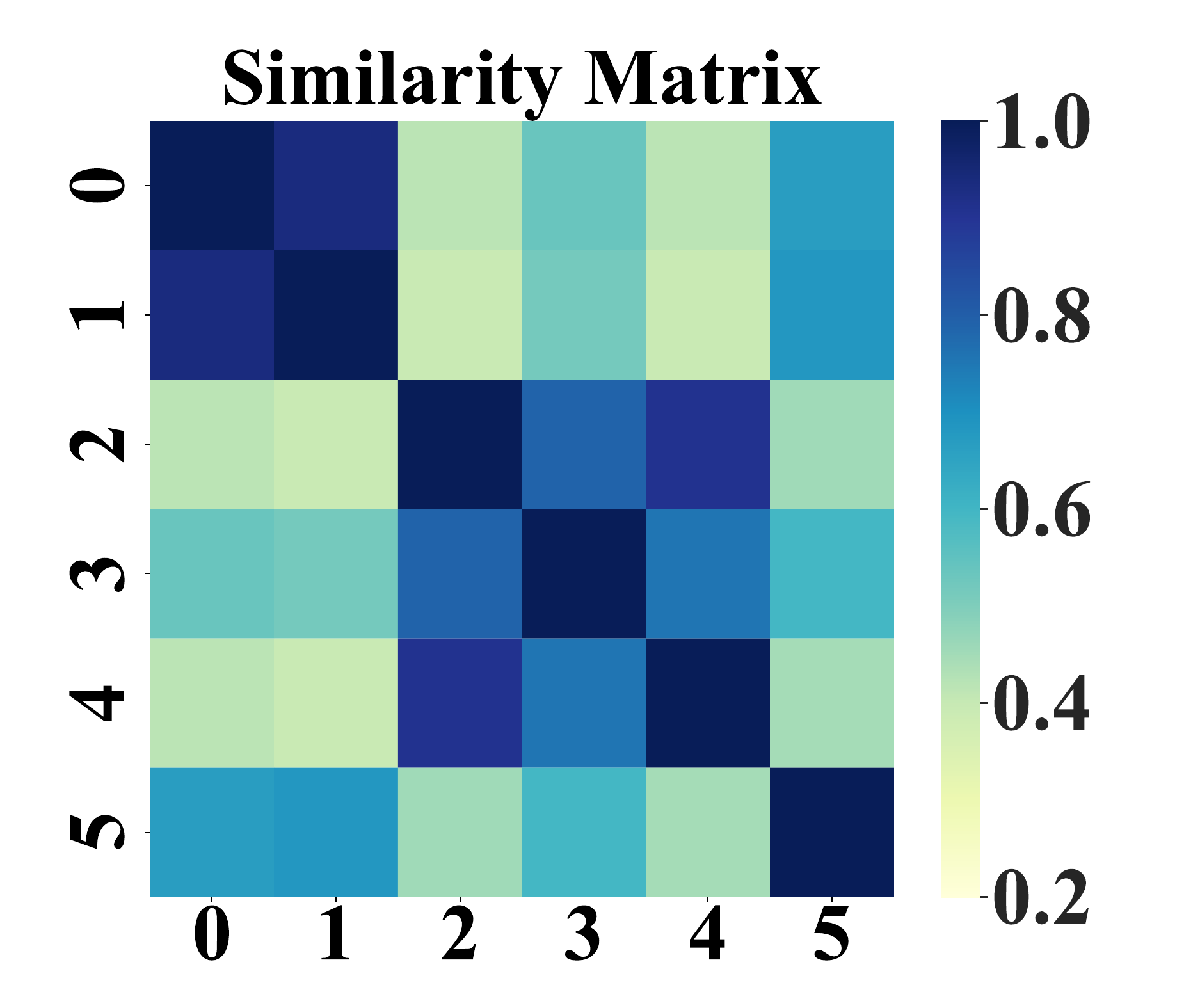}
    }
        \subfloat[]
    {
        \label{fusion-sim}
        \includegraphics[width=0.2\linewidth]{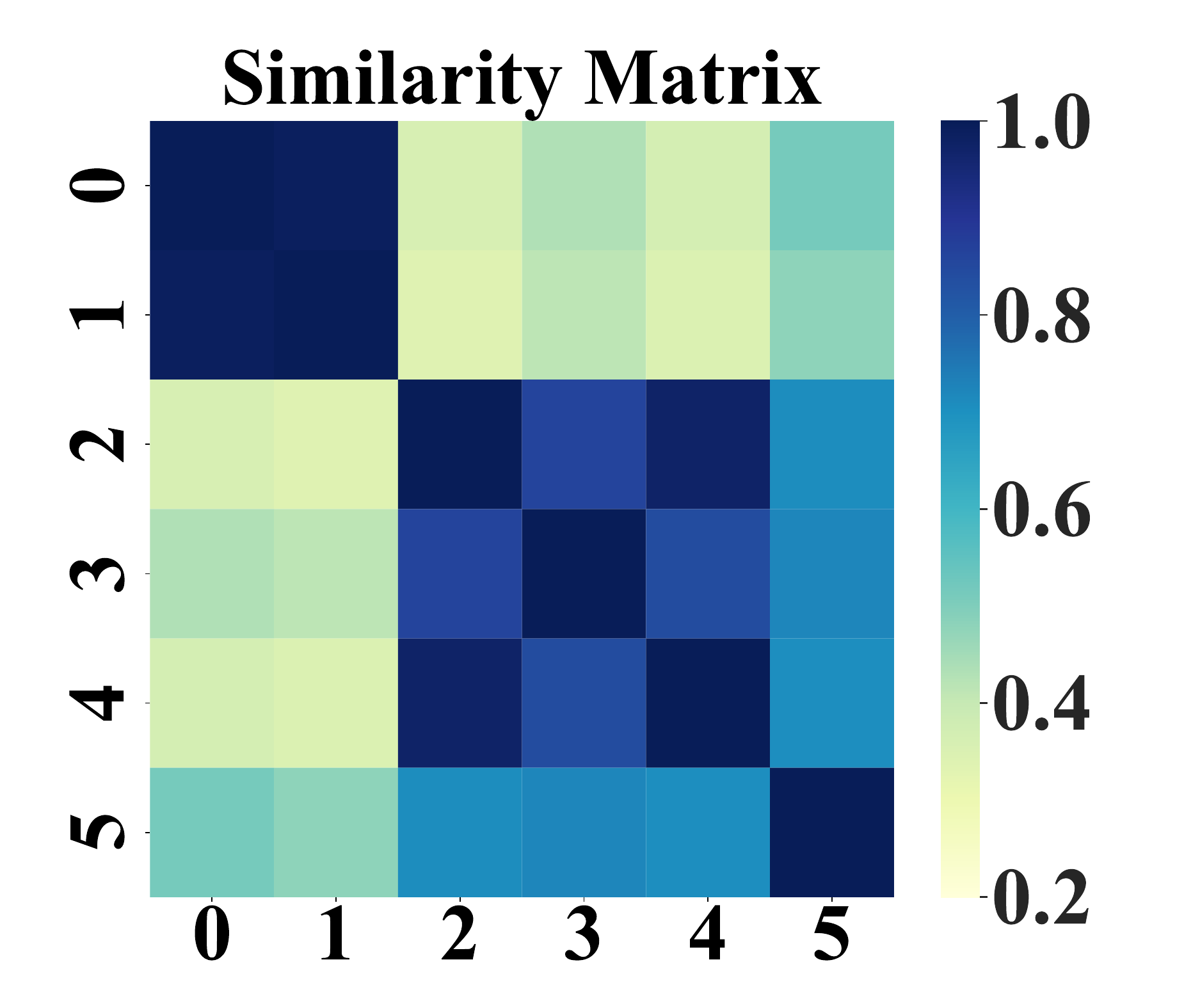}
    }
    \caption{(a): A subgraph of Cora dataset. The same classes of nodes are annotated in the same color, and the number denotes the node index.
    (b)-(e): The heat maps of similarity matrices in latent space with node embeddings obtained by graph partition (PAR) SSL task, node attributes clustering (CLU) SSL task, AUTOSSL, and dynamic fusion operation. 
    }
    \label{intro-demo}
\end{figure*}

Many studies on designing SSL methods in AGC have been proposed. Generative learning-based methods ~\cite{ARVGA, GraphMAE} aim to reconstruct graph structure or masked node features to learn node representations.  
Contrastive learning-based methods assist clustering by maximizing the mutual information of positive samples~\cite{dgi}. 
Predictive learning-based methods construct pretext tasks by exploiting natural supervision provided by the graph data itself~\cite{PAR}. 
Despite showing effectiveness, the advantage of using a single SSL task is limited due to ignoring helpful information from other tasks~\cite{AutoSSL}.
 
To address the problem, 
several methods are proposed to simultaneously learn multiple graph SSL tasks and show promising results.  
\cite{AutoSSL} proposed an AUTOSSL model that automatically searches the weights of SSL tasks in a multi-task learning framework by maximizing the designed pseudo-homophily to learn node embeddings. 
\cite{ParetoGNN} designed a PARETOGNN model that minimizes the potential conflicts between different SSL tasks to learn better node representations.

Although these methods can automatically learn the weights of various SSL tasks, they are globally shared in the whole graph. This leads to a quite straightforward concern: Is a fixed strategy of fusing different SSL tasks appropriate for learning good representations of all the graph nodes? 

We empirically explore and analyze the question on the Cora dataset. Fig. \ref{intro-demo}(a) displays a subgraph of Cora, in which nodes of the same classes are annotated in the same color. We visualize the similarity matrix of nodes obtained by different SSL tasks, \emph{i.e.,} node attributes clustering (CLU) and graph partition (PAR) SSL tasks \cite{PAR}, respectively, in Figs. \ref{intro-demo}(b) and (c).
CLU encourages the encoder to extract attribute information of nodes to assist the AGC task. In contrast, PAR emphasizes the learning of local topology information by the encoder. 
In Fig. \ref{intro-demo}, we observe that node representations learned from different SSL tasks show different similarity relationships between nodes, and both of them can not reflect the true clustering structure and lead to unsatisfactory clustering results. 

Furthermore, we use AUTOSSL to learn the node embeddings by automatically searching the weights of PAR and CLU tasks. The visualization of its corresponding similarity matrix is shown in Fig.\ref{intro-demo}(d). 
To learn proper weights for most nodes, AUTOSSL prefers to use the topology information extracted from the PAR task, making the distribution of the similarity matrix roughly the same as the similarity matrix obtained by the PAR task. Due to all nodes sharing a set of SSL task weights, AUTOSSL can not learn the intrinsic clustering structure. 

Intuitively, to obtain better clustering results, Nodes \#2-\#5 should use more attribute information, while Nodes \#0-\#1 should use more topology information. 
Therefore, we obtain the node-wise fusion representations by weighing two kinds of features and subsequently visualize the fusion representations. 
As shown in Fig. \ref{intro-demo}(e), it can clearly reflect the clustering structure. 
From this experiment, we empirically find that each node in the graph is different and it should adaptively fuse features extracted from various SSL tasks to learn discriminative node representations for clustering.

To tackle the above problem, we propose a novel framework, Dynamically Fusing Self-Supervised Learning (DyFSS), to fully utilize various graph self-supervised information for node-wise representation construction. 
Specifically, a dynamic fusion network based on the Mixture of Expert (MoE) framework is carefully designed for fusing features extracted from multiple SSL tasks. In this design, each SSL task is assigned to an expert to extract task-specific features. 
Each node can adaptively select a set of weights by a gating network and perform feature fusion. Such adaptability facilitates learning discriminative node representations. 
After that, we calculate the clustering distribution of fusion embeddings by using student's $t$-distribution to achieve end-to-end clustering. 
In the initial training stage, clustering alignment loss is unreliable owing to the subpar quality of fused features. To address this, we design a dual-level self-supervised strategy (\emph{i.e.,}, pseudo-label level, and graph structure level). 
This innovative strategy offers effective guidance for the fusion network, thereby enhancing node representations. 
Overall, the contributions of this paper are summarized as follows:
\begin{itemize}

\item We find the problem that the fusion strategy shared by all the nodes is not appropriate for learning node embeddings for clustering. Accordingly, we propose a multi-task SSL graph clustering framework that dynamically fuses the features extracted from multiple SSL tasks for each node using distinct weights derived from a gating network. 


\item To achieve reliable dynamic fusion, we design a dual-level supervised strategy to train the fusion network. The pseudo-label supervised information facilitates distinguishing different clusters, while the graph structure supervised information enhances the reliability of the fusion embeddings by capturing the underlying connections between nodes. 

\item Extensive experiment results on five public benchmark datasets verify that DyFSS can effectively boost the clustering performance by combing various SSL tasks. Our experiments show that DyFSS improves the accuracy metric by up to 8.66\% on the Photo dataset over the state-of-the-art multi-task SSL methods. 

\end{itemize}

\section{Related Work}
\subsection{Self-Supervised Learning in GNNs}

Graph Neural Networks (GNNs) possess robust representation capabilities in analyzing graph-structured data, while they rely on expensive task-specific labels. Many self-supervised learning methods are proposed to alleviate the demand for labeled data in graph-related tasks. 

Generative SSL techniques strive to reconstruct the adjacency matrix, masked node features, or both, as demonstrated in studies like \cite{ARVGA, GraphMAE}. Contrastive approaches \cite{dgi} leverage positive and negative samples to acquire node representations. Predictive methodologies, such as \cite{Pairdis, PAR}, formulate pretext tasks using inherent supervisory signals derived from the graph data.

Recently, \cite{AutoSSL} proposed AUTOSSL, and \cite{ParetoGNN} proposed PARETOGNN, have emerged as approaches that leverage multiple SSL tasks to enhance node representations and achieve improved generalization performance in downstream tasks. Both methods employ the multi-task learning framework, enabling automatic loss weight adjustment for each SSL task. 

Our method designs a dynamic fusion network, which automatically fuses the diversity features extracted from various SSL tasks according to the initial node embeddings of each node for clustering.

\begin{figure*}[!t]
    \centering
    \includegraphics[width=0.95\linewidth]{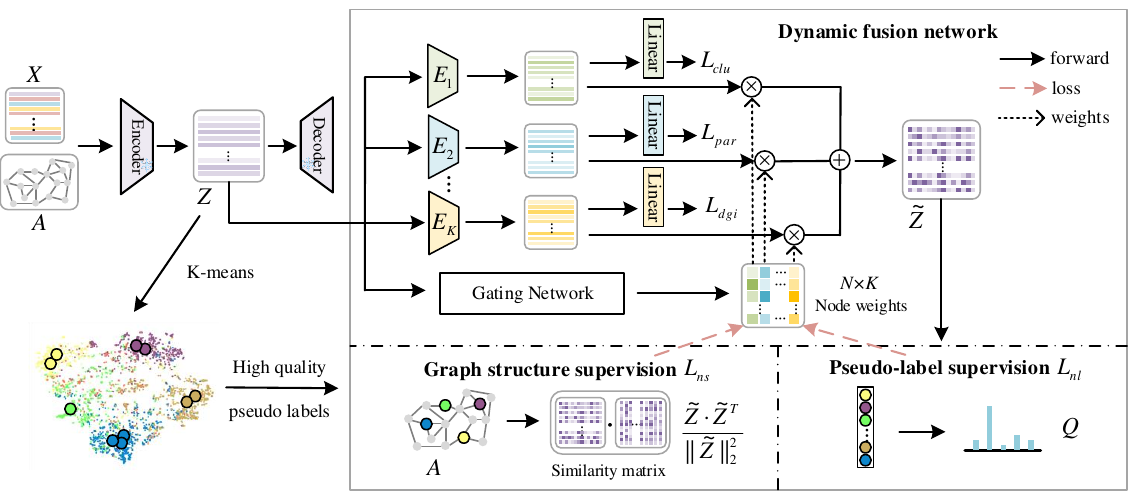}
    \caption{Architecture of DyFSS model. 
We first use the pre-trained model to obtain the initial node embeddings. The embeddings are then fed into the dynamic fusion network to obtain the fusion embeddings. 
Specifically, 
each SSL task is allocated to a task-specific GCN layer (\emph{i.e.,} an expert) to extract features using the corresponding SSL loss, \emph{i.e.,}$L_{clu}$, $L_{par}$, $L_{dgi}$, \emph{etc}. Simultaneously, the gating network generates a set of weights for each node, culminating in the subsequent execution of the feature fusion operation. 
Lastly, we use high-quality labels and graph structure as supervised information to provide effective guidance for the training of the dynamic fusion network.
    }
    \label{model}
\end{figure*}

\subsection{Attributed Graph Clustering}

Efficiently leveraging varied information inherent in graphs to learn distinctive node representations holds the utmost importance for the AGC task. Benefiting from the rapid development of GNNs, researchers have developed a series of AGC algorithms based on GNNs.

In particular, researchers proposed the graph auto-encoder \cite{VGAE} along with its variations \cite{ARVGA, RWR}. These methods aim to obtain node representations through graph data reconstruction, followed by the K-means algorithm to yield clustering results. Building upon this, the deep clustering network proposed by \cite{DAEGC} achieves simultaneously learn node representations and cluster assignments. 
Additionally, some methods \cite{SCAGC, CCGC} also use contrastive learning to guide the clustering process learning by comparing the similarities and differences between nodes. 

Differing from these approaches, DyFSS uses various information from multiple SSL tasks for clustering. Furthermore, we design a dual-level supervised strategy to enhance the node representations.

\section{Method}

\subsection{Notations and Overview}

Given an attributed graph $G'=\left\{V', E', \bm{X}\right\}$, where $V'=\left\{ v_1, v_2, ..., v_{N}\right\}$ representations the node set with $N$ nodes, $E'$ is the edge set of the graph, $\bm{X}=\left\{ \bm{x_1}, \bm{x_2}, ..., \bm{x_{N}}\right\} \in \mathbb{R}^{N \times d }$ is the initial feature matrix, and $\bm{x_i} = \bm{X}\left[i, : \right] \in \mathbb{R}^{d}$ denotes the node feature of $v_i$. The topological structure of graph $G'$ is represented by an adjacency matrix $\bm{A} \in \mathbb{R}^{N \times N}$, where $a_{ij}=\bm{A}\left[i,j\right]=1$ if $\left(v_i, v_j\right) \in E'$, otherwise $a_{ij}=0$. 
To keep the scale of the feature vectors unchanged, adjacency matrix $\bm{A}$ needs to be normalized through calculating \bm{$\widetilde{D}^{-\frac{1}{2}}\widetilde{A}\widetilde{D}^{-\frac{1}{2}}$}, where $\bm{\widetilde{A}}=\bm{A}+\bm{I}$ is the adjacency matrix with self-connection, $\bm{\widetilde{D}}=\sum_{j}\tilde{a}_{ij}$ is the degree matrix of $\bm{\widetilde{A}}$, and $\bm{I}$ is the identity matrix. 

In this section, we will introduce the novel AGC method named Dynamically Fusion Self-Supervised Learning (DyFSS), which aims to adaptively fuse the features extracted from multiple SSL tasks for clustering. 
As illustrated in Fig. \ref{model}, DyFSS mainly consists of two components, \emph{i.e.,} a dynamic fusion network and a dual-level of self-supervised strategy module. 
Next, we will introduce the proposed two modules and the final network objectives in detail. 

\subsection{Dynamic Fusion Network}

Achieving an adaptive node-wise fusion of features from diverse SSL tasks becomes crucial to enhance node representations for clustering.  
The existing multi-task learning frameworks lack the ability to achieve dynamic node-wise fusion. 
The Mixture of Experts (MoE) \cite{MOE} network is an ensemble learning technique, which can address this limitation.

Therefore, we propose the dynamic fusion network, leveraging the MoE framework to fuse the features extracted from multiple SSL tasks for each node. 
Specifically, the dynamic fusion network comprises $K$ expert networks $E_1,..., E_K$, alongside a gating network $G$ that produces a $K$-dimensional vector. In this design, each expert is assigned to an individual SSL task, responsible for extracting task-specific features.

\subsubsection{Expert Network.}

Each expert network is a GCN layer that aggregates information from neighbors of each node. For the $k^{th}$ expert, the output $\bm{Z_k} \in \mathbb{R}^{N \times d'}$ is formulated as
\begin{equation}
\label{get_z_k}
    \bm{Z_{k}} = f_{linear}(\bm{Z}, \bm{A}| \bm{W^{(k)}}) = \phi(\bm{\widetilde{D}^{-\frac{1}{2}}\widetilde{A}\widetilde{D}^{-\frac{1}{2}}ZW^{(k)}}),
\end{equation}
where $\bm{Z} \in \mathbb{R}^{N \times d'}$ is the latent node representations obtained by the pre-trained model, and \bm{$W^{(k)}$} is the parameter to be trained. 
For node $v_i$, the output of $k^{th}$ expert is represented as $E_k(\bm{z_i})=\bm{Z_k}[i, :] \in \mathbb{R}^{d'}$.

Given the presence of $K$ SSL tasks, each expert is designated to an individual SSL task, functioning as an encoder responsible for extracting task-specific features to facilitate fusion.
The graph encoder of $k^{th}$ SSL task is represented as $f_{\theta_g^{(k)}}(\cdot; \theta_g^{(k)})$, where the $\theta_g^{(k)} = \bm{W^{(k)}}$, and it is obvious that  $f_{\theta_g^{(k)}}(G') = \bm{Z_k}$. 
The loss function $k^{th}$ task is represented as $L_k(G';\theta_g^{(k)}, \theta_t^{(k)})$, where the $\theta_t^{(k)}$ refers to the task-specific parameters (\emph{i.e.,} a linear layer). 
Then, we minimize the total loss of all SSL tasks, enabling the extraction of task-specific features for each expert,
\begin{equation}
    L_{ssl} = \sum_{k=1}^{K} L_k(G';\theta_g^{(k)}, \theta_t^{(k)}).
\end{equation}

\subsubsection{Gating Network.}
Utilizing the gating network, we derive a set of expert weights for each node based on its corresponding initial node embedding, and it is formulated as:
\begin{equation}
\label{get_gating}
    G(\bm{z_i}) = softmax(\bm{z_i} \cdot \bm{W_{n}}),
\end{equation}
where the $\bm{z_i}=\bm{Z}[i,:]$ is the initial node embeddings of node $v_i$, \bm{$W_{n}$} is the trainable parameter, and $softmax(\cdot)$ is a function that converts input vector into weights for each SSL expert. 

\subsubsection{Node-wise Feature Fusion.}

For the node $v_i$, with the output of each expert $E_1(\bm{z_i}),..., E_K(\bm{z_i})$ and the output of the gating network $G(\bm{z_i})$, a linear combination operation is used to fuse the features from different experts,
\begin{equation}
\label{get_fusion_emb}
    \bm{\tilde{z}_i} = \sum_{k=1}^{K} G(\bm{z_i})_k E_k(\bm{z_i}),
\end{equation}
where the $G(\bm{z_i})_k$ represents the $k^{th}$ element of $G(\bm{z_i})$, and \bm{$\tilde{z}_i$} is the fused embedding of node $v_i$. Finally, we can obtain the fused node representations $\bm{\widetilde{Z}} \in  \mathbb{R}^{N \times d'}$ of all nodes for clustering.

\subsection{Dual-Level Self-Supervised Strategy}



To train the dynamic fusion network, a conventional approach offers guidance by aligning the distribution between the soft clustering distribution and target distribution \cite{DAEGC, SDCN}. 
However, owing to the subpar quality of fused features in the initial training stage, the clustering alignment loss is unreliable, potentially leading to optimization in the wrong direction. 

To address this issue, we design a dual-level self-supervised strategy to train the dynamic fusion network. Some simple samples can learn better embeddings through pretraining, and high-quality pseudo labels generated by these nodes are used as supervised information to distinguish distinct clusters and enhance the compactness of fusion representations within each cluster. 
In addition, the learned initial node representations may be inaccurate for other difficult samples. 
Therefore, we use the original adjacency structure of the graph itself as supervised information, which ensures that the embedding representation captures the true underlying connections between nodes, enhancing the reliability of the fused  node representations. 


\subsubsection{Pseudo-Label Level Self-Supervised Strategy.}
After obtaining latent node representations \bm{$Z$}, followed by \cite{DAEGC, CDRS}, the student's t-distribution is used to calculate the clustering assignment distribution $\bm{Q} \in \mathbb{R}^{N \times C}$. Subsequently, the pseudo-label node set $M$ and the corresponding pseudo-label set $Y$ are generated as:

\begin{equation}
\label{get_node_set}
    M=\left\{ i \, | \, (\max_{j\in \left\{1,..., C \right\}} q_{ij} )\geq \gamma, \forall i\in V\right\},
\end{equation}

\begin{equation}
\label{get_pseudo_labels}
    Y=\left\{ \mathop{\mathrm{argmax}}\limits_{j\in \left\{1,..., C \right\}} \, q_{ij} \,| \, \forall i\in V \right\},
\end{equation}
where the $\gamma \in (0, 1)$ represents the $m$-th percentile $P_m$ calculated from the maximum clustering scores across all nodes in \bm{$Q$}.

Next, we calculate the soft clustering distribution $\bm{\widetilde{Q}} \in \mathbb{R}^{N \times C}$ of fusion embeddings $\bm{\widetilde{Z}}$. 
After that, 
the cross-entropy loss is used to achieve the pseudo-label level supervision of nodes, formulated as:
\begin{equation}
    L_{nl} = - \frac{1}{|M|} \sum_{i=1}^{|M|} \sum_{j=1}^{C}\tilde{y}'_{ij}log(\tilde{q}_{ij}),
\end{equation}
where the $\tilde{y}'_{ij}$ is the one-hot pseudo label of node $v_i \in M$ after alignment between different clustering results from \bm{$Q$} and \bm{$\widetilde{Q}$}.

\subsubsection{Graph Structure Level Self-Supervised Strategy.}

Specifically, after obtaining the fusion embeddings \bm{$\widetilde{Z}$}, we scale the embeddings to the $[0, 1]$ interval by min-max scaler for variance reduction. Subsequently, we utilize the cosine function to calculate the pair-wise similarity matrix $\bm{\widetilde{S}} \in \mathbb{R}^{N \times N}$ of nodes, formulated as:
\begin{equation}
\label{eq-s}
    \bm{\widetilde{S}} = \frac{\bm{\widetilde{Z}\widetilde{Z}^T}}{||\bm{\widetilde{Z}}||_{2}^{2}}.
\end{equation}
Then we formulate the graph structure supervision by employing the cross-entropy loss, as follows:
\begin{equation}
\label{eq-ns}
    L_{ns} = \frac{1}{N^2} \sum_{i=1}^{N}\sum_{j=1}^{N} -\tilde{a}_{ij} log(\tilde{s}_{ij})-(1-\tilde{a}_{ij})log(1-\tilde{s}_{ij}),
\end{equation}
where $\tilde{a}_{ij}$ is the element of the adjacency matrix with self-connection \bm{$\widetilde{A}$}. 

\begin{table*}[t!]
   \renewcommand\tabcolsep{3.5pt}     
 \renewcommand\arraystretch{1}        
 \small
\begin{tabular}{c|c|ccc|ccc|ccc|ccc|ccc}
   \toprule
   \multirow{2}{*}{Dataset} & \multirow{2}{*}{Type}     
  & \multicolumn{3}{c|}{Cora} & \multicolumn{3}{c|}{Citeseer} & \multicolumn{3}{c|}{Photo} & \multicolumn{3}{c|}{Computers} & \multicolumn{3}{c}{CS}\\   
  \cmidrule{3-17}
  & & ACC & NMI & F1 & ACC & NMI & F1 & ACC & NMI & F1 & ACC & NMI & F1 & ACC & NMI & F1 \\
   \midrule
   K-Means & - & 33.64 & 13.76 & 23.39 & 58.04 & 31.64 & 55.23 & 40.55 & 28.92 & 24.37 & 34.36 & 17.41 & 13.10  & 58.13 & 64.27 & 39.68\\
   AE & S-F & 43.06 & 24.15 & 42.84 & 53.65 & 26.43 & 50.44 & 42.65 & 35.51 & 40.36  & 31.39 & 31.30 & 25.79  &  55.14 & 56.35 & 37.51\\
   DeepWalk & S-G & 54.74 & 31.82 & 54.48 & 33.63 & 12.13 & 33.47 & \underline{76.14} & 66.72 & 70.26 &  50.68 & 49.37 & \textbf{47.75} &  54.69 & 54.51 &  52.70\\
   \midrule
   VGAE & S-F\&G & 61.67 & 46.10 & 57.67 & 54.19 & 28.64 & 52.43 & 54.00 & 49.49 & 53.18 &  44.28 & 39.48 & 37.77  & 61.67 & 69.09 & 55.49\\
   ARVGA & S-F\&G & 63.80 & 45.00 & 62.70 & 54.40 & 26.10 & 52.90 & 43.24 & 32.77 & 34.84 & 28.27 & 26.81 & 18.62 & 64.44 & 71.62 & 59.03\\
   DAEGC & S-F\&G & 67.02 & 48.22 & 66.19 & 62.73 & 37.82 & 60.64 & 62.27 & 58.88 & 60.35 & 52.46 & 46.38 & 42.22 & 68.65 & 71.23 & 54.31\\
   GraphMAE & S-F\&G & 62.78 & 54.75 & 63.74 & 68.95 & 43.74 & 64.25 & 75.42 & \underline{67.25} & 69.01 & \underline{53.04} & \underline{52.66} & 40.76 & 59.83 & 70.85 & 49.27\\
   DGI & S-F\&G & \underline{69.90} & \underline{54.95} & 66.95 & 69.13 & 44.48 & 64.36 & 44.07 & 36.74 & 36.56 & 39.96 & 35.11 & 25.02 & 69.45 & 76.97 & 66.27 \\
   GCA & S-F\&G & 60.86 & 48.16 & 57.94 & 66.97 & 41.53 & 63.25 & 63.19 & 51.80 & 52.10 & 49.43 & 43.48 & 38.95 & 66.98 & 74.34 & 64.87 \\
   CCA-SSG & S-F\&G & 64.73 & 54.63 & 54.85 & 67.93 & 42.89 & 61.04 & 69.59 & 57.88 & 66.21 & 47.59 & 44.56 & 42.69 & \underline{73.31} & 73.64 & 58.98\\
   BGRL & S-F\&G & 69.28 & 54.91 & 67.15 & 64.20 & 38.40 & 59.78 & 75.01 & 65.68 & \underline{71.41} & 46.94 & 44.12 & 41.45 & 69.82 & 76.46 & \underline{71.52} \\
   \midrule
   AUTOSSL & M-F\&G & 69.57 & 52.82 & \underline{67.21} & \underline{69.16} & \underline{44.59} & \underline{64.37} & 68.17 & 52.76 & 63.31 & 50.58 & 44.62 & 37.36  & 68.92 & \textbf{76.98} & 66.17\\
   PARETOGNN & M-F\&G & 68.55 & 52.17 & 59.17 & 68.27 & 43.28 & 61.30 & 71.15 & 64.02 & 66.77 & 47.18 & 47.09 & 46.69 & 69.82 & 70.49 & 65.34\\
   DyFSS & M-F\&G & \textbf{72.19} & \textbf{55.49}& \textbf{68.09} & \textbf{70.18} & \textbf{44.80} & \textbf{64.68} & \textbf{79.81} & \textbf{71.86} & \textbf{73.56} & \textbf{56.95} & \textbf{53.68} & \underline{46.78} & \textbf{75.81} & \underline{76.80} & \textbf{72.69} \\
   \bottomrule
\end{tabular}
\centering
 \caption{Clustering performance on five datasets. The bold and underlined values indicate the best and the runner-up results, respectively. The notation S-F represents models that leverage single SSL task with node features as inputs. The notation M-F\&G represents models that concurrently exploit both node features and graph structure as inputs and utilize varied information from multiple SSL tasks.}
 \label{baselines-smalldata}%
\end{table*}


The overall learning objective comprises four parts, \emph{i.e.,} the task-specific loss of all SSL tasks, the pseudo-label level loss, the graph structure level loss, and the clustering loss:
\begin{equation}
\label{get_total_loss}
    L = L_{nl} + L_{ns} + \lambda_{1}L_{ssl} + \lambda_{2}L_{pq},
\end{equation}
where $\lambda_1$ and $\lambda_2$ are hyper-parameters that balance the weight of different losses. $L_{pq}$ is the clustering alignment loss \cite{DAEGC, pq-clustering1, pq-clustering2}. 
The detailed learning procedure of our method is shown in Algorithm \ref{our:algorithm}.

\begin{algorithm}[b!] 
\caption{DyFSS}
\label{our:algorithm}
\textbf{Input}:Graph $G'=(\bm{X}, \bm{A})$. Pseudo-label node set $M$ and pseudo-label set $Y$. Initial node embeddings $\bm{Z}$. Iteration number $I$. Hyper-parameters $\lambda_1, \lambda_2$.  Number of classes $C$.\\
\textbf{Output}: Clustering results $R$. 
\begin{algorithmic}[1] 
\STATE Generate the labels of five SSL tasks.
\STATE Initialize the parameters of $K$ experts and the gating network to obtain $\bm{\widetilde{Z}}$.
\STATE Initialize the clustering centers and clustering results $O$ with K-means based on $\bm{\widetilde{Z}}$.
\STATE Align the pseudo labels set $Y$ with the initial clustering results $O$.
\FOR{$i=1, 2, ..., I$}
\STATE Update $\bm{\widetilde{Z}}$ by Eq.(\ref{get_z_k}), Eq.(\ref{get_gating}) and Eq.(\ref{get_fusion_emb}).
\STATE Calculate soft assignment distributions \bm{$\widetilde{Q}$} and pairwise similarity matrix \bm{$\widetilde{S}$} by Eq.(\ref{eq-s}).
\STATE Calculate $L_{nl}$, $L_{ns}$, $L_{ssl}$, and $L_{pq}$ respectively.
\STATE Update the whole network by minimizing  Eq.(\ref{get_total_loss}). 
\ENDFOR
\STATE Obtain the clustering results $R$ by final soft clustering distribution \bm{$\widetilde{Q}$}.
\STATE \textbf{return} $R$.
\end{algorithmic}
\end{algorithm}

\section{Experiments}
\subsection{Experiment Setup}
\subsubsection{Benchmark Datasets and Evaluation Metrics.}

We conducted experiments on five widely used real-world datasets in the literature \cite{AutoSSL,cca-ssg}, including Cora, Citeseer, Photo, Computers, and CS. 
Three widely used metrics were employed to evaluate the clustering results: Accuracy (ACC), Normalized Mutual information (NMI), and macro F1-score (F1). 

\begin{figure*}[!t]
     \centering
        \includegraphics[width=1\linewidth]{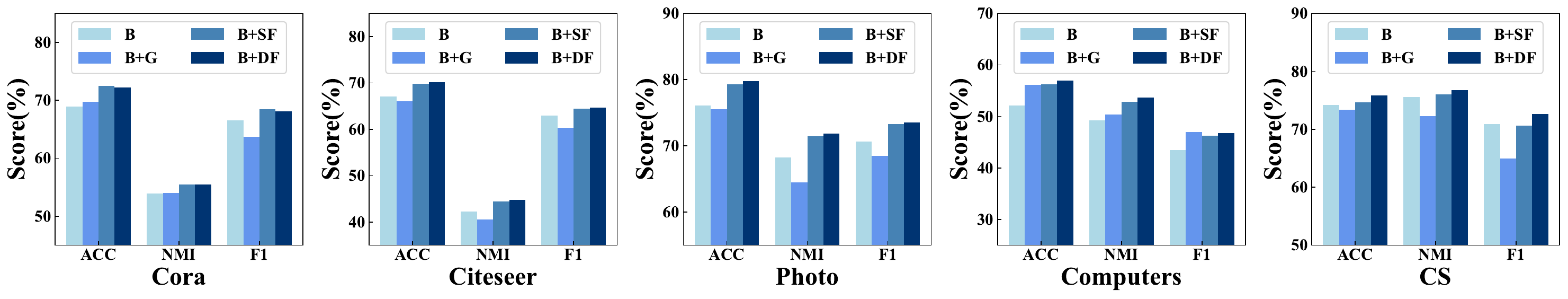}
    \caption{Ablation studies of dynamic fusion network. The baseline (B) is a per-trained ARVGA model, and B+DF is our method.
    The B+G indicates the baseline involving dynamic gating mechanisms and fixed experts. Conversely, B+SF signifies the baseline that uses a fixed gating network in combination with dynamic experts, representing a form of static fusion approach.
    }
    \label{ablFusion}
\end{figure*}

\subsubsection{Baseline Methods.}

We compared our proposed method with thirteen baselines to evaluate the effectiveness of our method. 
K-means \cite{k-means} and AE proposed by \cite{AE} are methods that only use node features for clustering, while DeepWalk proposed by \cite{DeepWalk} only uses the graph structure to learn node embeddings. 
Additionally, some GCN-based baseline methods use both node attributes and graph structure to learn node representations. 
VGAE \cite{VGAE}, ARVGA \cite{ARVGA}, DAEGC \cite{DAEGC}, and GraphMAE \cite{GraphMAE} are generative methods. DGI \cite{dgi}, GCA \cite{GCA}, CCA-SSG \cite{cca-ssg}, and BGRL \cite{BGRL} are contrastive methods. 
Furthermore, we compared the clustering results with the state-of-the-art multi-task self-supervised learning methods, \emph{i.e.,} AUTOSSL \cite{AutoSSL} and ParetoGNN \cite{ParetoGNN}.

\subsubsection{Basic SSL task and Training Procedure.}
 
For SSL tasks, we chose the five SSL pretext tasks adopted by AUTOSSL, including PAR, CLU, PAIRDIS, PAIRSIM, and DGI. 
The training of DyFSS includes two stages. Firstly, we pre-trained the base model to obtain node representations \bm{$Z$} for at least 100 epochs with the multi-task framework by minimizing reconstruction loss and self-supervised loss of five pretext tasks. 
Thereafter, we trained the fusion clustering network with node representations \bm{$Z$} under the guidance of Eq.(\ref{get_total_loss}) for at least 200 epochs until convergence. 
Our method was implemented with the PyTorch platform, an NVIDIA 1070 GPU, and an NVIDIA 3090 GPU.

\subsubsection{Parameters Setting.}

For PARETOGNN, we reproduced their model by using the same five SSL tasks as our method and following the parameter settings of the original paper. For AUTOSSL, we use the AUTOSSL-DS strategy to save the search cost. 
For other baselines, we reproduced their source code by following the setting of the original literature and reported corresponding clustering results. 
For our method, we used the ARVGA as the base model, and the encoder of it applied a two-layer GCN with 256 and 128 filters. All ablation studies were trained with the Adam optimizer, and the learning rate was set to 1e-3. For trade-off hyper-parameters, we set $\lambda_2$ to 1. Furthermore, $\lambda_1$ was set to 0.01 for Photo and Computers datasets and 0.1 for others. Moreover, $m$ was set to 40 for the Citeseer dataset and 50 for others. 

\subsection{Clustering Performance Comparisons}

The comparative clustering performance of our method and thirteen baselines across five benchmark datasets is outlined in Tab. \ref{baselines-smalldata}. 
It can be observed from these results that:

\textbf{(1) The node-wise dynamic fusion operation boosts the clustering performance compared to the multi-task SSL learning methods.}
Different from AUTOSSL and PARETOGNN, DyFSS achieves node-wise dynamic fusion operation by integrating features from various SSL tasks. Compared with these methods, DyFSS obtains remarkable performances. 
For example, DyFSS outperforms AUTOSSL by 2.62\%, 2.67\%, and 0.88\% on the Cora dataset, with respect to ACC, NMI, and F1, respectively. 
The experiments show the effectiveness of DyFSS, underscoring its efficacy in enhancing feature representations.

\textbf{(2) Fully utilizing various information from multiple SSL tasks facilitates the clustering task.}
Compared with the methods that only use a single SSL task, the methods utilizing multiple SSL tasks demonstrate a significant performance advantage. 
Additionally, for different datasets, the contribution of various information to the AGC task varies. We observe that DeepWalk outperforms AE by 33.49\% on Photo, while AE outperforms DeepWalk by 20.02\% on Citeseer, in terms of ACC. 
It shows that the Photo dataset prefers structure information while Citeseer prefers attribute information for clustering. 
Therefore, it is necessary to utilize various information from SSL tasks to assist in clustering for different datasets.

\textbf{(3) Algorithms that incorporate both feature and structure information tend to yield superior clustering results compared to methods that only leverage one source of information.}
Across the majority of datasets, GCN-based methods exhibit superior performance compared with AE and DeepWalk.


\begin{table}[!t]
\centering
\small
   \renewcommand\tabcolsep{7pt}     
 \renewcommand\arraystretch{0.9}        
\begin{tabular}{c|c|c|c|c}
   \toprule
   \multicolumn{1}{c|}{Datasets} & \multicolumn{1}{c|}{Model} & ACC & NMI & F1\\
   \midrule
   \multirow{3}{*}{Cora} & DyFSS w/o $L_{dual}$ & 70.75 & 55.40 & 66.99 \\ 
                         & DyFSS w/o $L_{ns}$ & 71.34 & 55.04 & 67.28   \\
                         & DyFSS & \textbf{72.19} & \textbf{55.49} & \textbf{68.09}  \\
   \midrule
    \multirow{3}{*}{Citeseer}& DyFSS w/o $L_{dual}$ & 68.53 & 43.73 & 64.19 \\ 
                             & DyFSS w/o $L_{ns}$ & 69.37 & 43.92 & 63.90   \\ 
                             & DyFSS & \textbf{70.18} & \textbf{44.80} & \textbf{64.68}   \\
   \midrule
    \multirow{3}{*}{Photo} & DyFSS w/o $L_{dual}$ & 77.83 & 70.28 & 72.72 \\ 
                           & DyFSS w/o $L_{ns}$ & 78.36 & 70.19 & 72.41   \\ 
                           & DyFSS & \textbf{79.81} & \textbf{71.86} & \textbf{73.56}   \\
    \midrule
    \multirow{3}{*}{Computers} & DyFSS w/o $L_{dual}$ & 47.96 & 52.18 & 45.14 \\ 
                               & DyFSS w/o $L_{ns}$ & 55.86 & 52.24 & 45.71   \\ 
                               & DyFSS  & \textbf{56.95} & \textbf{53.68} & \textbf{46.78}   \\
   \midrule
   \multirow{3}{*}{CS} & DyFSS w/o $L_{dual}$  & 75.56 & \textbf{76.81} & 72.52 \\ 
                       & DyFSS w/o $L_{ns}$ & 75.60 & 76.39 & 71.63   \\ 
                       & DyFSS & \textbf{75.81} & 76.80 & \textbf{72.69}   \\
   \bottomrule
\end{tabular}
\caption{Ablation comparisons of the dual-level supervised strategy $L_{dual}$ which includes pseudo-label level supervised information $L_{nl}$ and graph structure level supervised information $L_{ns}$.
}
\label{ablation}
\end{table}

\begin{figure*}[!t]
    \centering
    \subfloat[Raw data]
    {
        \includegraphics[width=0.15\linewidth]{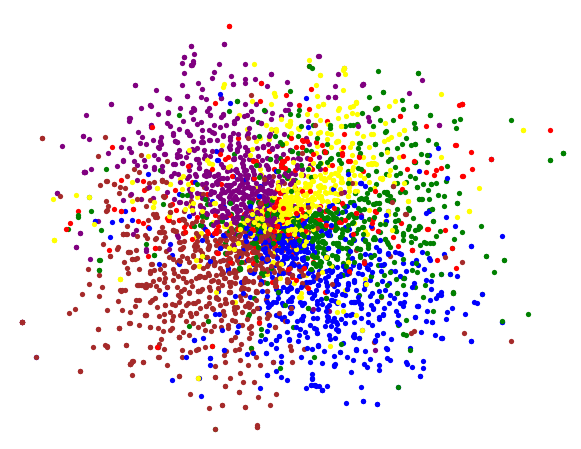}
    }
    \subfloat[VGAE]
    {
        \includegraphics[width=0.15\linewidth]{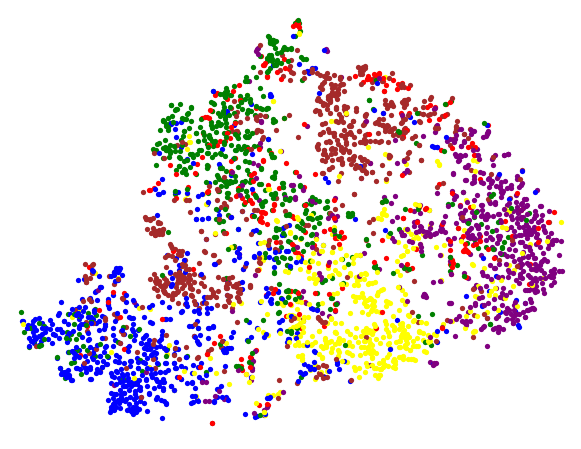}
    } 
    \subfloat[DGI]
    {
        \includegraphics[width=0.15\linewidth]{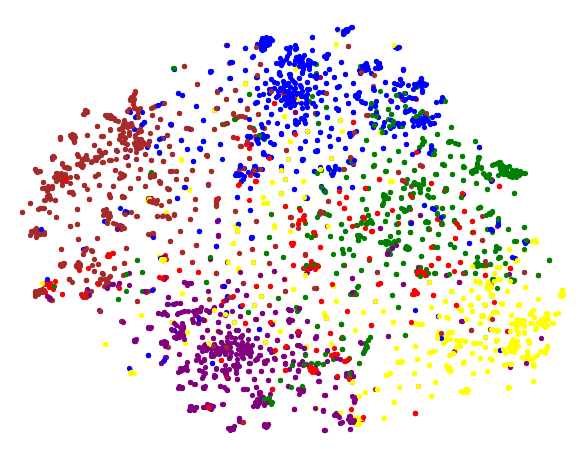}
    }
     \subfloat[GraphMAE]
    {
        \includegraphics[width=0.15\linewidth]{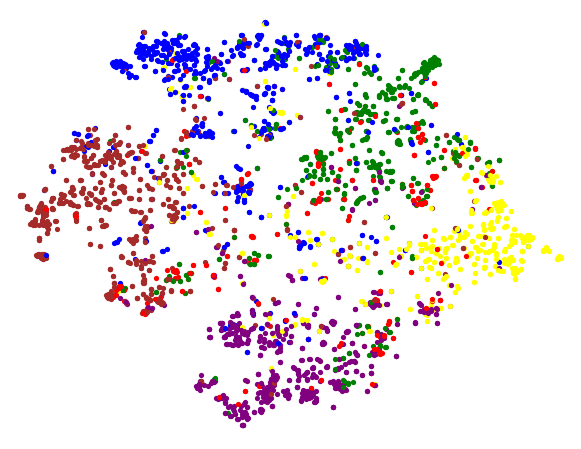}
    }
     \subfloat[AUTOSSL]
    {
        \includegraphics[width=0.15\linewidth]{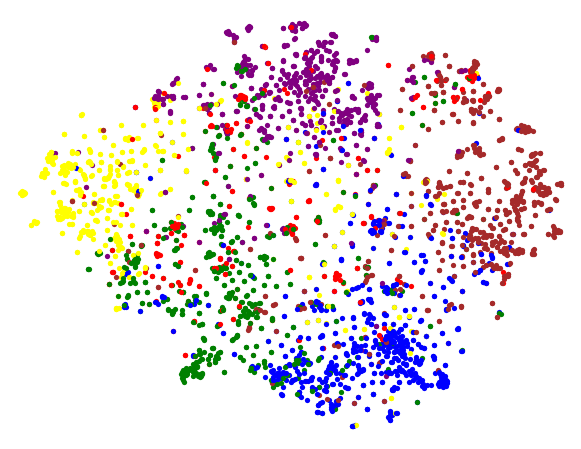}
    }
    \subfloat[DyFSS]
    {
        \includegraphics[width=0.15\linewidth]{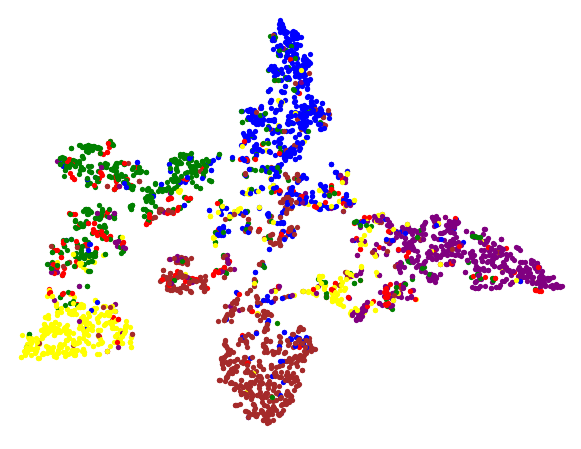}
    }
    \caption{$t$-SNE visualization on Citeseer dataset. }
    \label{tnse}
\end{figure*}

\begin{figure}[!t]
    \centering
    \includegraphics[width=0.7\linewidth]{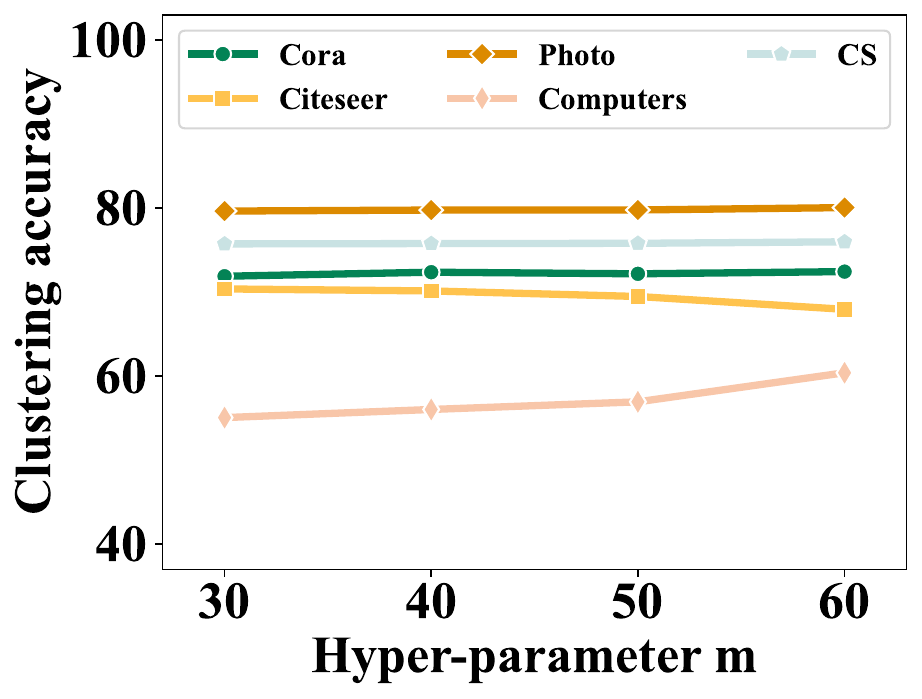}
    \caption{The sensitivity analysis of DyFSS with variation of hyper-parameter $m$ on five datasets.}
    \label{hyper-param}
\end{figure}

\begin{figure}[!t]
    \centering
    \includegraphics[width=0.85\linewidth]{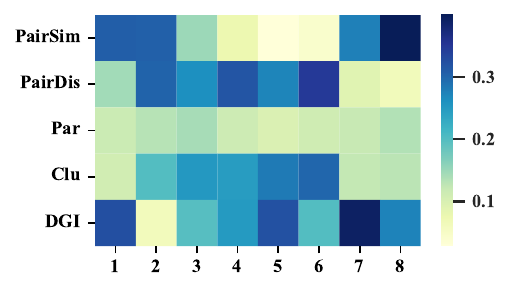}
    \caption{Visualization of the weights assigned by five SSL experts on the Citeseer dataset. The x-axis shows eight randomly selected nodes and the y-axis presents five SSL tasks.}
    \label{node-weights}
\end{figure}

\subsection{Ablation Studies}


\subsubsection{Effectiveness of Dynamically Fusion Network.}
In this part, we conducted ablation studies to verify the effectiveness of the dynamic fusion network and reported the results in Fig. \ref{ablFusion}. 
From the figures, we can observe that B+DF and B+SF consistently outperform the B and B+G methods, which shows that fixed experts are not conducive to clustering and it is necessary to learn the multiple features for different experts. 
Additionally, the B+DF method improves the clustering performance over the B+SF method on four of five datasets. 
Both of the observations collectively demonstrate that the novel dynamic fusion network possesses the capability to extract a range of diverse features and dynamically leverage these features to generate node representations, leading to improved clustering decisions.

\subsubsection{Effectiveness of Dual-Level Supervised Strategy.} 
To verify the superiority of the dual-level supervised strategy, we compared our method with the other two counterparts. 
The results of these methods are summarized in Tab. \ref{ablation}.
In terms of ACC, DyFSS w/o $L_{ns}$ outperforms the DyFSS w/o $L_{dual}$ by 0.84\%, 0.53\%, and 7.90\% on Citeseer, Photo, and Computers datasets, respectively. 
Additionally, DyFSS demonstrates even further advancements, 
surpassing the DyFSS w/o $L_{ns}$ method by 0.88\%, 1.67\%, and 1.44\% on the same respective datasets, in terms of NMI.  
Based on the results, it is evident that the proposed dual-level supervised strategy avoids the problem of clustering alignment loss deviating from the optimal optimization direction, and each kind of supervised information contributes to the final performance. 
Notably, the pseudo-label supervised information aids in distinguishing diverse clusters and enhancing the compactness of node representations within each cluster. 
Concurrently, the guidance from graph structure supervision complements the pseudo-label supervised information, 
facilitating learning reliable fusion embeddings. 

\subsection{Further Analysis}

\subsubsection{Hyper-Parameter Analysis of $m$.}
 
We investigate the influence of hyper-parameter $m$, which plays a pivotal role in establishing the threshold $\gamma$ for the generation of pseudo labels. 
We conduct experiments to show the effect of this parameter on all datasets. 
Based on the results illustrated in Fig. \ref{hyper-param}, it can be observed that DyFSS is insensitive to hyper-parameter $m$. 
Except for the Citeseer dataset, the accuracy metric has slight increments with the increasing value $m$. 
Notably, our method demonstrates stable performance across a wide range of $m$.

\subsubsection{Visualization.}

For an intuitive demonstration of the effectiveness of our method, we employ the $t$-SNE algorithm \cite{tsne} on the Citeseer dataset. This allows us to visualize the distribution of the raw data, and the learned node embeddings produced by four baseline methods, alongside our DyFSS model, in a two-dimensional space. 
As shown in Fig. \ref{tnse}, the visual results clearly demonstrate DyFSS's enhanced capability in revealing the inherent clustering structure within the data and improving the compactness of learned node representations within clusters.

Furthermore, we present a visualization of the weights assigned to five SSL tasks involving different nodes. Illustrated in Fig. \ref{node-weights}, diverse nodes obtain varying sets of weights, underscoring the ability of the dynamic fusion network to generate a set of weights for each node flexibly.

\section{Conclusion}

To solve the problem that all the nodes share the same set of SSL weights, we proposed a novel Dynamically Fusing Self-Supervised Learning (DyFSS) method for the attributed graph clustering task. 
This method enables every node within the graph to dynamically fuse distinct features derived from various SSL tasks, yielding discriminative node representations. 
Additionally, the designed dual-level supervised strategy provides effective guidance for the training of the proposed dynamic fusion network, which facilitates distinguishing different clusters and enhancing the compactness of node embeddings within each cluster. 
Experiments on five benchmark datasets corroborate the effectiveness of our method in boosting the clustering performance. 
Furthermore, 
our method is insensitive to the hyper-parameter and the learned node representations can reflect the intrinsic clustering structure.

\section*{Acknowledgments}

This work was supported in part by the National Key R\&D Program of China under Grant 2022ZD0116500 and in part by the National Natural Science Foundation of China under Grants 62106174, 62222608, 62266035, 61925602, U23B2049, and 62076179.

\bibliography{aaai24}

\begin{thebibliography}{26}
\providecommand{\natexlab}[1]{#1}

\bibitem[{Bo et~al.(2020)Bo, Wang, Shi, Zhu, Lu, and Cui}]{SDCN}
Bo, D.; Wang, X.; Shi, C.; Zhu, M.; Lu, E.; and Cui, P. 2020.
\newblock Structural Deep Clustering Network.
\newblock In \emph{{WWW} '20: The Web Conference 2020, Taipei, Taiwan, April 20-24, 2020}, 1400--1410. {ACM} / {IW3C2}.

\bibitem[{Guo et~al.(2017)Guo, Gao, Liu, and Yin}]{pq-clustering2}
Guo, X.; Gao, L.; Liu, X.; and Yin, J. 2017.
\newblock Improved Deep Embedded Clustering with Local Structure Preservation.
\newblock In \emph{Proceedings of the Twenty-Sixth International Joint Conference on Artificial Intelligence, {IJCAI} 2017, Melbourne, Australia, August 19-25, 2017}, 1753--1759. ijcai.org.

\bibitem[{Hartigan and Wong(1979)}]{k-means}
Hartigan, J.~A.; and Wong, M.~A. 1979.
\newblock A K-Means Clustering Algorithm.
\newblock \emph{Journal of the Royal Statistical Society: Series C (Applied Statistics)}, 28(1): 100--108.

\bibitem[{Hou et~al.(2022)Hou, Liu, Cen, Dong, Yang, Wang, and Tang}]{GraphMAE}
Hou, Z.; Liu, X.; Cen, Y.; Dong, Y.; Yang, H.; Wang, C.; and Tang, J. 2022.
\newblock GraphMAE: Self-Supervised Masked Graph Autoencoders.
\newblock In \emph{{KDD} '22: The 28th {ACM} {SIGKDD} Conference on Knowledge Discovery and Data Mining, Washington, DC, USA, August 14 - 18, 2022}, 594--604. {ACM}.

\bibitem[{Jin et~al.(2022)Jin, Liu, Zhao, Ma, Shah, and Tang}]{AutoSSL}
Jin, W.; Liu, X.; Zhao, X.; Ma, Y.; Shah, N.; and Tang, J. 2022.
\newblock Automated Self-Supervised Learning for Graphs.
\newblock In \emph{The Tenth International Conference on Learning Representations, {ICLR} 2022, Virtual Event, April 25-29, 2022}, 1--13. OpenReview.net.

\bibitem[{Ju et~al.(2023)Ju, Zhao, Wen, Yu, Shah, Ye, and Zhang}]{ParetoGNN}
Ju, M.; Zhao, T.; Wen, Q.; Yu, W.; Shah, N.; Ye, Y.; and Zhang, C. 2023.
\newblock Multi-task Self-supervised Graph Neural Networks Enable Stronger Task Generalization.
\newblock In \emph{The Eleventh International Conference on Learning Representations, {ICLR} 2023, Kigali, Rwanda, May 1-5, 2023}, 1--13. OpenReview.net.

\bibitem[{Kipf and Welling(2016)}]{VGAE}
Kipf, T.~N.; and Welling, M. 2016.
\newblock Variational Graph Auto-Encoders.
\newblock \emph{CoRR}, abs/1611.07308.

\bibitem[{Mrabah et~al.(2023)Mrabah, Bouguessa, Touati, and Ksantini}]{RGAE}
Mrabah, N.; Bouguessa, M.; Touati, M.~F.; and Ksantini, R. 2023.
\newblock Rethinking Graph Auto-Encoder Models for Attributed Graph Clustering (Extended abstract).
\newblock In \emph{39th {IEEE} International Conference on Data Engineering, {ICDE} 2023, Anaheim, CA, USA, April 3-7, 2023}, 3891--3892. {IEEE}.

\bibitem[{Pan et~al.(2018)Pan, Hu, Long, Jiang, Yao, and Zhang}]{ARVGA}
Pan, S.; Hu, R.; Long, G.; Jiang, J.; Yao, L.; and Zhang, C. 2018.
\newblock Adversarially Regularized Graph Autoencoder for Graph Embedding.
\newblock In \emph{Proceedings of the Twenty-Seventh International Joint Conference on Artificial Intelligence}.

\bibitem[{Peng et~al.(2020)Peng, Dong, Luo, Wu, and Zheng}]{Pairdis}
Peng, Z.; Dong, Y.; Luo, M.; Wu, X.; and Zheng, Q. 2020.
\newblock Self-Supervised Graph Representation Learning via Global Context Prediction.
\newblock \emph{CoRR}, abs/2003.01604.

\bibitem[{Perozzi, Al{-}Rfou, and Skiena(2014)}]{DeepWalk}
Perozzi, B.; Al{-}Rfou, R.; and Skiena, S. 2014.
\newblock DeepWalk: online learning of social representations.
\newblock In \emph{The 20th {ACM} {SIGKDD} International Conference on Knowledge Discovery and Data Mining, {KDD} '14, New York, NY, {USA} - August 24 - 27, 2014}, 701--710. {ACM}.

\bibitem[{Shazeer et~al.(2017)Shazeer, Mirhoseini, Maziarz, Davis, Le, Hinton, and Dean}]{MOE}
Shazeer, N.; Mirhoseini, A.; Maziarz, K.; Davis, A.; Le, Q.~V.; Hinton, G.~E.; and Dean, J. 2017.
\newblock Outrageously Large Neural Networks: The Sparsely-Gated Mixture-of-Experts Layer.
\newblock In \emph{5th International Conference on Learning Representations, {ICLR} 2017, Toulon, France, April 24-26, 2017, Conference Track Proceedings}. OpenReview.net.

\bibitem[{Thakoor et~al.(2022)Thakoor, Tallec, Azar, Azabou, Dyer, Munos, Velickovic, and Valko}]{BGRL}
Thakoor, S.; Tallec, C.; Azar, M.~G.; Azabou, M.; Dyer, E.~L.; Munos, R.; Velickovic, P.; and Valko, M. 2022.
\newblock Large-Scale Representation Learning on Graphs via Bootstrapping.
\newblock In \emph{The Tenth International Conference on Learning Representations, {ICLR} 2022, Virtual Event, April 25-29, 2022}. OpenReview.net.

\bibitem[{Vaibhav, Huang, and Frederking(2019)}]{RWR}
Vaibhav; Huang, P.; and Frederking, R.~E. 2019.
\newblock {RWR-GAE:} Random Walk Regularization for Graph Auto Encoders.
\newblock \emph{CoRR}, abs/1908.04003.

\bibitem[{van~der Maaten and Hinton(2008)}]{tsne}
van~der Maaten, L.; and Hinton, G. 2008.
\newblock Visualizing Data using t-SNE.
\newblock \emph{Journal of Machine Learning Research}, 9(86): 2579--2605.

\bibitem[{Velickovic et~al.(2019)Velickovic, Fedus, Hamilton, Li{\`{o}}, Bengio, and Hjelm}]{dgi}
Velickovic, P.; Fedus, W.; Hamilton, W.~L.; Li{\`{o}}, P.; Bengio, Y.; and Hjelm, R.~D. 2019.
\newblock Deep Graph Infomax.
\newblock In \emph{7th International Conference on Learning Representations, {ICLR} 2019, New Orleans, LA, USA, May 6-9, 2019}, 1--13. OpenReview.net.

\bibitem[{Wang et~al.(2019)Wang, Pan, Hu, Long, Jiang, and Zhang}]{DAEGC}
Wang, C.; Pan, S.; Hu, R.; Long, G.; Jiang, J.; and Zhang, C. 2019.
\newblock Attributed Graph Clustering: {A} Deep Attentional Embedding Approach.
\newblock In \emph{Proceedings of the Twenty-Eighth International Joint Conference on Artificial Intelligence, {IJCAI} 2019, Macao, China, August 10-16, 2019}, 3670--3676. ijcai.org.

\bibitem[{Wu et~al.(2023)Wu, Lin, Tan, Gao, and Li}]{SSLGNNs:Survey}
Wu, L.; Lin, H.; Tan, C.; Gao, Z.; and Li, S.~Z. 2023.
\newblock Self-Supervised Learning on Graphs: Contrastive, Generative, or Predictive.
\newblock \emph{{IEEE} Trans. Knowl. Data Eng.}, 35(4): 4216--4235.

\bibitem[{Xia et~al.(2022)Xia, Wang, Gao, Yang, and Gao}]{SCAGC}
Xia, W.; Wang, Q.; Gao, Q.; Yang, M.; and Gao, X. 2022.
\newblock Self-consistent Contrastive Attributed Graph Clustering with Pseudo-label Prompt.
\newblock \emph{IEEE Transactions on Multimedia}, 1--13.

\bibitem[{Xie, Girshick, and Farhadi(2016)}]{pq-clustering1}
Xie, J.; Girshick, R.~B.; and Farhadi, A. 2016.
\newblock Unsupervised Deep Embedding for Clustering Analysis.
\newblock In \emph{Proceedings of the 33nd International Conference on Machine Learning, {ICML} 2016, New York City, NY, USA, June 19-24, 2016}, volume~48 of \emph{{JMLR} Workshop and Conference Proceedings}, 478--487. JMLR.org.

\bibitem[{Yang et~al.(2017)Yang, Fu, Sidiropoulos, and Hong}]{AE}
Yang, B.; Fu, X.; Sidiropoulos, N.~D.; and Hong, M. 2017.
\newblock Towards K-means-friendly Spaces: Simultaneous Deep Learning and Clustering.
\newblock In \emph{Proceedings of the 34th International Conference on Machine Learning, {ICML} 2017, Sydney, NSW, Australia, 6-11 August 2017}, volume~70 of \emph{Proceedings of Machine Learning Research}, 3861--3870. {PMLR}.

\bibitem[{Yang et~al.(2023)Yang, Liu, Zhou, Wang, Tu, Zheng, Liu, Fang, and Zhu}]{CCGC}
Yang, X.; Liu, Y.; Zhou, S.; Wang, S.; Tu, W.; Zheng, Q.; Liu, X.; Fang, L.; and Zhu, E. 2023.
\newblock Cluster-Guided Contrastive Graph Clustering Network.
\newblock In \emph{Thirty-Seventh {AAAI} Conference on Artificial Intelligence, {AAAI} 2023, Thirty-Fifth Conference on Innovative Applications of Artificial Intelligence, {IAAI} 2023, Thirteenth Symposium on Educational Advances in Artificial Intelligence, {EAAI} 2023, Washington, DC, USA, February 7-14, 2023}, 10834--10842. {AAAI} Press.

\bibitem[{You et~al.(2020)You, Chen, Wang, and Shen}]{PAR}
You, Y.; Chen, T.; Wang, Z.; and Shen, Y. 2020.
\newblock When Does Self-Supervision Help Graph Convolutional Networks?
\newblock In \emph{Proceedings of the 37th International Conference on Machine Learning, {ICML} 2020, 13-18 July 2020, Virtual Event}, volume 119 of \emph{Proceedings of Machine Learning Research}, 10871--10880. {PMLR}.

\bibitem[{Zhang et~al.(2021)Zhang, Wu, Yan, Wipf, and Yu}]{cca-ssg}
Zhang, H.; Wu, Q.; Yan, J.; Wipf, D.; and Yu, P.~S. 2021.
\newblock From Canonical Correlation Analysis to Self-supervised Graph Neural Networks.
\newblock In \emph{Advances in Neural Information Processing Systems 34: Annual Conference on Neural Information Processing Systems 2021, NeurIPS 2021, December 6-14, 2021, virtual}, 76--89.

\bibitem[{Zhu et~al.(2022)Zhu, Li, Wang, Xiao, Zhao, and Hu}]{CDRS}
Zhu, P.; Li, J.; Wang, Y.; Xiao, B.; Zhao, S.; and Hu, Q. 2022.
\newblock Collaborative Decision-Reinforced Self-Supervision for Attributed Graph Clustering.
\newblock \emph{IEEE Transactions on Neural Networks and Learning Systems}, 1--13.

\bibitem[{Zhu et~al.(2021)Zhu, Xu, Yu, Liu, Wu, and Wang}]{GCA}
Zhu, Y.; Xu, Y.; Yu, F.; Liu, Q.; Wu, S.; and Wang, L. 2021.
\newblock Graph Contrastive Learning with Adaptive Augmentation.
\newblock In \emph{{WWW} '21: The Web Conference 2021, Virtual Event / Ljubljana, Slovenia, April 19-23, 2021}, 2069--2080. {ACM} / {IW3C2}.

\end{thebibliography}

\end{document}